\DeclareUnicodeCharacter{266B}{\textmusicalnote}

\documentclass[sigconf]{acmart}
\usepackage{mdframed}

\AtBeginDocument{%
  }

\setcopyright{acmlicensed}
\copyrightyear{2024}
\acmYear{2024}
\acmDOI{}
\acmISBN{}

\acmConference[KDDCup '24]{2024 KDD Cup Workshop for Retrieval Augmented Generation at the 30th ACM SIGKDD Conference on Knowledge Discovery and Data Mining}{Aug 25--29,
  2024}{Barcelona, Spain}

\begin{document}

\title{Honest AI: Fine-Tuning "Small" Language Models to Say "I Don't Know", and Reducing Hallucination in RAG}


\author{Xinxi Chen}
\authornote{These authors contributed equally to this work. Emails are provided for contact purposes only and do not represent an endorsement by any institution.}

\affiliation{%
  \institution{Independent Researcher}
  }
\email{xc336@cornell.edu}

\author{Li Wang}
\authornotemark[1]
\affiliation{%
  \institution{Independent Researcher}}
\email{li@liwang.info}

\author{Wei Wu}
\authornotemark[1]
\affiliation{%
  \institution{Independent Researcher}}
 \email{drweiwu@outlook.com}

\author{Qi Tang}
\authornotemark[1]
\affiliation{%
  \institution{Independent Researcher}}
\email{qitang@closeby-ai.com}

\author{Yiyao Liu}
\authornotemark[1]
\affiliation{%
  \institution{Independent Researcher}}
\email{info@yiyaoliu.com}

\renewcommand{\shortauthors}{Chen et al.}
\begin{abstract}
  Hallucination is a key roadblock for applications of Large Language Models (LLMs), particularly for enterprise applications that are sensitive to information accuracy. To address this issue, two general approaches have been explored: Retrieval-Augmented Generation (RAG) to supply LLMs with updated information as context, and fine-tuning the LLMs with new information and desired output styles. In this paper, we propose Honest AI: a novel strategy to fine-tune "small" language models to say "I don't know" to reduce hallucination, along with several alternative RAG approaches. The solution ranked 1st in Task 2 for the false premise question\footnotemark[1]. The alternative approaches include using RAG with search engine and knowledge graph results, fine-tuning base LLMs with new information and combinations of both approaches. Although all approaches improve the performance of the LLMs, RAG alone does not significantly improve the performance and fine-tuning is needed for better results. Finally, the hybrid approach achieved the highest score in the CRAG benchmark \cite{yang2024cragcomprehensiverag}. In addition, our approach emphasizes the use of relatively small models with fewer than 10 billion parameters, promoting resource efficiency.
    \footnotetext[1]{The Meta CRAG Challenge has over 2000 participants with over 5500 submissions.}
\end{abstract}

\ccsdesc[500]{Computing methodologies~Natural language generation}

\keywords{Large Language Models (LLM), Retrieval Augmented Generation (RAG), Knowledge Graph, Search}


\maketitle

\section{Introduction}
Large Language Models (LLMs), as a type of foundation models with general language capabilities, have eclipsed traditional Natural Language Processing (NLP) models that focus on specific tasks in majority of NLP applications since the inception of GPT 
\cite{brown2020language}. Supervised fine tuning (SFT) using 
labeled data and reinforcement learning from human feedback 
(RLHF) using preference data have proven effective in further enhancing LLMs' performance and alignment (e.g., ChatGPT \cite{ouyang2022training}  for question answering applications). However, LLMs suffer from hallucination, which hinders their application in accuracy-sensitive scenarios, such as the enterprise applications. 

\begin{figure}[h]
  \centering
  \includegraphics[width=\linewidth]{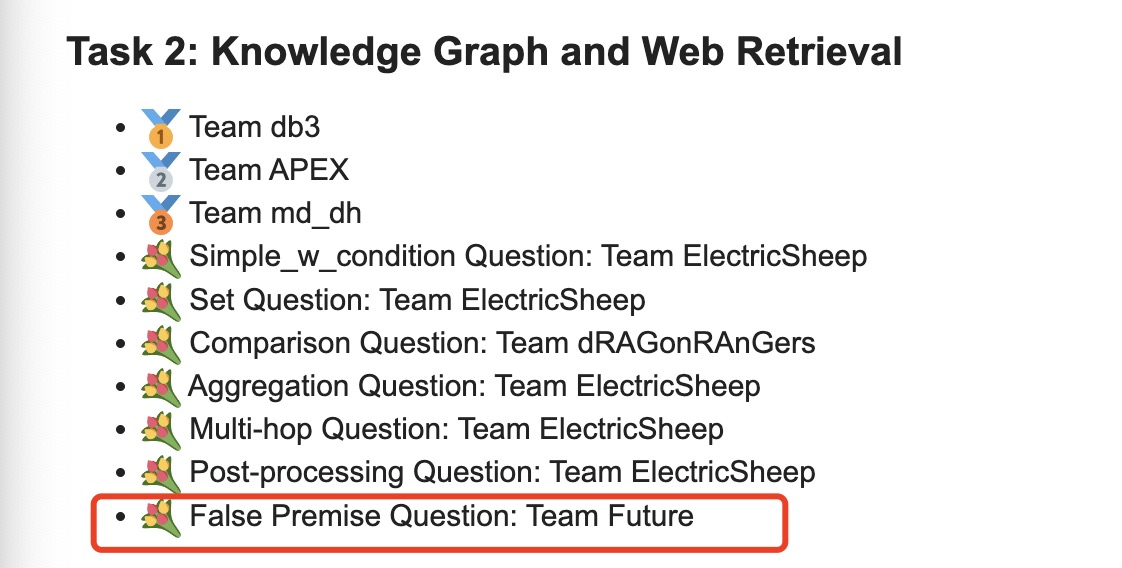}
  \caption{Team Future in Meta CRAG Challenge 2024 Winner List}
  \label{fig:ranking}
\end{figure}

To alleviate the hallucinations of LLMs, several approaches have been proposed, including retrieval-augmented generations (RAGs) \cite{lewis2021retrievalaugmentedgenerationknowledgeintensivenlp} and fine-tuning with domain-specific knowledge. In this paper, we summarize the approaches we tried in the 2024 Meta KDD Cup competition with Comprehensive RAG Benchmark (CRAG) data \cite{yang2024cragcomprehensiverag}. Since the CRAG benchmark focuses on difficult problems for LLMs, vanilla LLMs fail to perform well right out of the box. It turns out that RAG alone is not enough to alleviate hallucination in the benchmark and fine-tuning is needed to achieve higher accuracy. Our results show that the hybrid approach using both RAG and fine tuning performs best in CRAG. With those approaches combined, our team, Team Future, ranked high in each task of the competition and won first place in the false premise question in Task 2 (Fig. ~\ref{fig:ranking}).

The review of related work is summarized in section 2, followed by a description of methodologies in section 3. The results are shown in section 4, with conclusions and future works discussed in section 5 and 6. 

\section{Related Work}
LLMs are very good at memorizing the content used for their pretraining. However, there are many 
drawbacks to using the memorization capability directly for Question Answering (QA) tasks. 
For example, depending on the 
size of the model, the quality of the pretraining data, and the type 
of questions, LLMs' memorization capability can be very limited and difficult to control. LLMs
are also challenging to update except through retraining or fine-tuning, so they cannot handle questions on recent events if deployed. Arguably, the most
problematic drawback of using LLMs for QA is that they can hallucinate, especially when the models
are unsure if the context contains the information needed to answer a question. 

Fine-tuning is a straightforward way to update the knowledge of LLMs when new information becomes available. However, due to the scarcity of GPUs and  limited access to high-quality data, it is not feasible for many use cases. Furthermore, the behavior of fine-tuning language models is not well studied, and fine-tuning with poor data or practices might decrease the model's capabilities, causing modality collapse unless more complex methods such as RLHF are used \cite{ouyang2022training}.

Retrieval-Augmented Generation (RAG) \cite{lewis2021retrievalaugmentedgenerationknowledgeintensivenlp, gao2024retrievalaugmentedgenerationlargelanguage} 
is a popular method to address the shortcomings of LLMs, by augmenting them with 
non-parametric data sources, and leveraging LLMs' powerful in-context learning \cite{NEURIPS2020_1457c0d6} 
capability. \cite{gao2024retrievalaugmentedgenerationlargelanguage} grouped 
the approaches of using RAG for LLMs into three categories: Naive RAG, 
Advanced RAG \cite{ma-etal-2023-query, zheng2024take}, and Modular RAG \cite{yu2023generate,shao-etal-2023-enhancing}. 
Naive and Advanced RAG approaches are widely used in 
practice due their simplicity and low development cost. These approaches generally 
consist of three parts: curating non-parametric databases, retrieving relevant snippets 
from the databases given the query, and generating responses using LLMs through in-context learning and prompt engineering with the related snippets. 

While the research on combining LLMs and RAG for QA mainly focuses on text, there
is also research exploring the use of resources beyond text, such as images \cite{chen2022muragmultimodalretrievalaugmentedgenerator}, audio \cite{zhao-etal-2023-generating}, video \cite{Yang2023Vid2SeqLP}, and code \cite{10172590} to enhance the capabilities of language models.

There are various efforts to create RAG benchmarks and proposing 
appropriate evaluation metrics in recent years. \cite{yang2024cragcomprehensiverag} is one 
of the recent ones, which forms the foundation of this paper. \cite{yang2024cragcomprehensiverag}
created a factual question answering benchmark of 4,409 question-answer pairs and 
mock APIs to simulate web and Knowledge Graph (KG) searches. It also proposes an evaluation mechanism that distinguishes between hallucinations and missing answers, and assigns a higher penalty to hallucinations. \cite{Chen_Lin_Han_Sun_2024} created a RAG evaluation benchmark in both 
English and Chinese, and analyzed different LLMs from 4 aspects: noise robustness,
negative rejection, information integration, and counterfactual. They found 
that LLMs demonstrate a certain degree of noise robustness, but struggle significantly
in other aspects. 

Apart from specific RAG datasets, there are many existing QA datasets that include context passages for each question. 
These datasets can also be used for RAG experiments, 
and cover a wide range of questions such as multiple-choice QA 
\cite{pang-etal-2022-quality,allenai:arc,talmor-etal-2019-commonsenseqa}, 
single-hop QA \cite{10.1162/tacl_a_00276,joshi-etal-2017-triviaqa,rajpurkar-etal-2016-squad},
multi-hop QA \cite{yang-etal-2018-hotpotqa,ho-etal-2020-constructing,trivedi-etal-2022-musique},
and domain-specific QA \cite{dasigi-etal-2021-dataset,moller-etal-2020-covid,wang-etal-2024-cmb}.

\section{Methodologies}

In this section, we will first introduce the CRAG dataset and then describe the main approaches we have tried, including native RAG, fine-tuning, and hybrid approaches. 

\subsection{Dataset}

The CRAG dataset includes five question domains (e.g. movie, sports, and etc.) with varying levels of complexity, ranging from straightforward facts to those requiring reasoning (e.g. false premise and multi-hop reasoning) \cite{yang2024cragcomprehensiverag}. It also considers facts with different levels of timeliness (e.g. real-time, fast-changing, slow-changing, and stable). Figure 2 illustrates the distribution of question types across different domains and timeliness categories. Notably, the characteristics of question distributions in the movie and finance domains differ significantly. Real-time and fast-changing questions necessitate access to relevant, up-to-date data sources and effective RAG implementations. In contrast, the movie domain contains more static questions than the finance domain, which may be easier to address.

\begin{figure}[h]
  \centering
  \includegraphics[width=\linewidth]{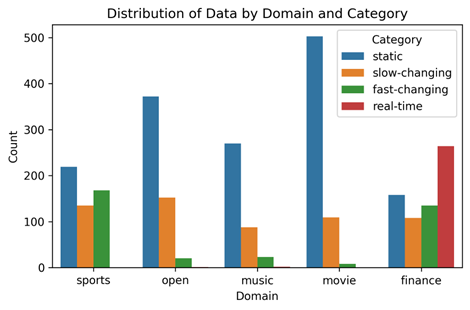}
  \caption{The Distribution of Questions in Different Domains and Types of Timeliness }
\end{figure}

\begin{figure*}[h]
  \centering
  \includegraphics[width=\linewidth]{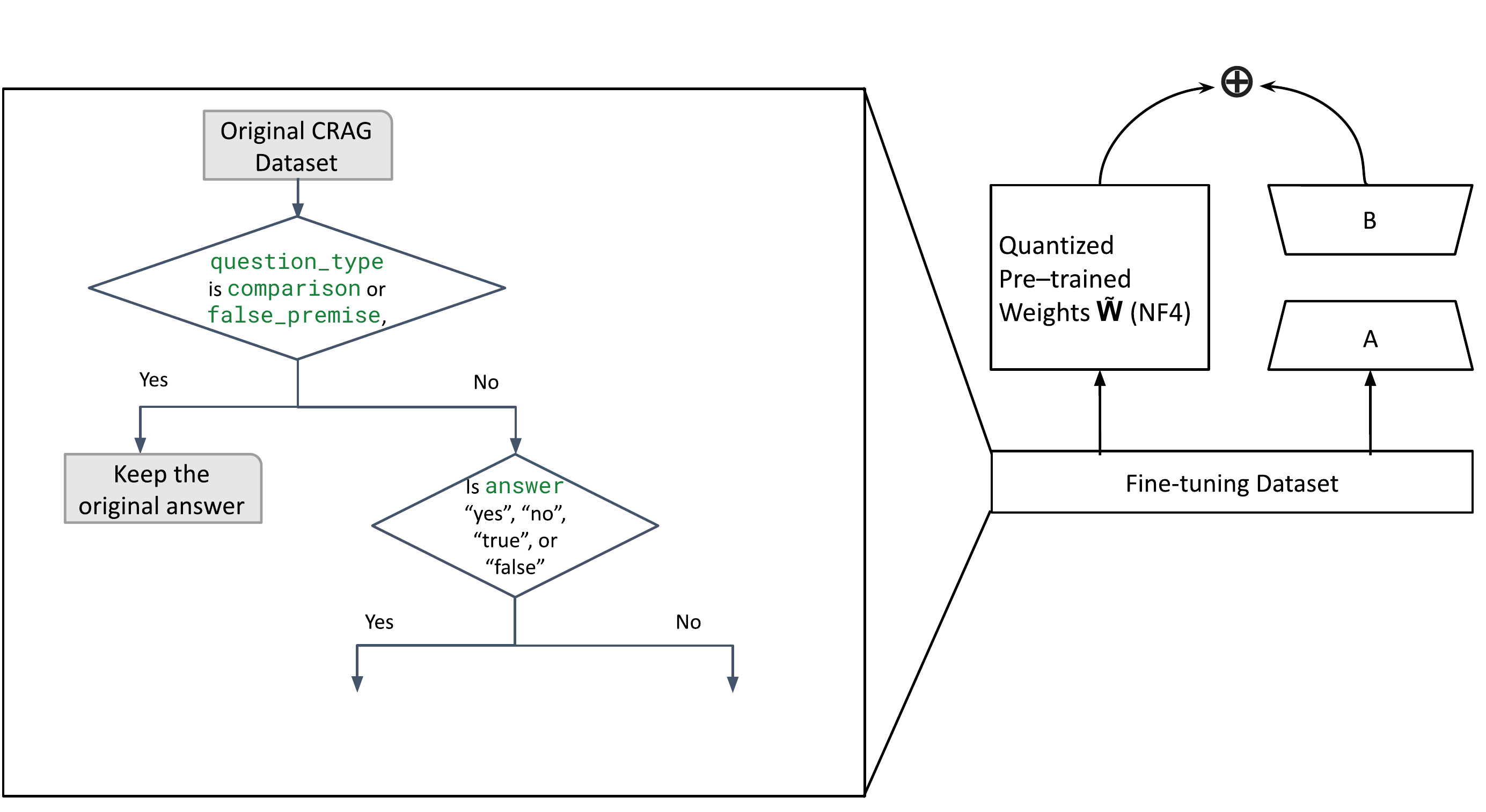}
  \caption{Fine-tuning Process}
  \label{fig:fine_tuning}
\end{figure*}

Our solution ranked first in Task 2 for false premise questions. A false premise question is defined as a question with a false assumption. For example, "What's the name of Taylor Swift's rap album before she transitioned to pop?" This is a false premise question because Taylor Swift didn't release any rap albums, and the expected answer is "invalid question" \cite{yang2024cragcomprehensiverag}.

In this competition, we chose a divide-and-conquer approach to design a RAG system to tackle different types of questions. We assumed that categorizing these questions could be done easily in practice. The objective is to identify the key parametric differences, such as varying prompts, parsing techniques, chunking sizes, top-k thresholds, and rule designs, to optimize performance and achieve a higher score during implementation.

\subsection{Scoring}
The overall score is a macro-average across all domains. The scoring system awards scores based on the quality of the response, and penalizes hallucination \cite{yang2024cragcomprehensiverag}. If the response is perfect, it receives 1 point. If the response is acceptable, it receives 0.5 points. If the response is missing, i.e. "I don't know", it receives 0 points. If the response is incorrect, it receives -1 point.

\subsection{RAG}

RAG \cite{lewis2021retrievalaugmentedgenerationknowledgeintensivenlp, gao2024retrievalaugmentedgenerationlargelanguage} is a popular approach to alleviate the hallucinations of LLMs. There are various architectures for RAG in real applications, including advanced RAG techniques. In this paper, we explored several naive RAG approaches. Generally, naive RAG selects the highest cosine similarity results from a vector database and supplies the context as inputs for LLMs. There are multiple ways to find the relevant information to implement RAG. We tried using naive cosine similarity and using LLMs as retrieval and ranking models to find relevant information for each web page.

Furthermore, as an extreme case, we also tried using the state-of-the-art Gemini 1.5 pro model with a 1 million token context window \cite{reid2024gemini} to supply all the retrieved web pages to LLMs. These models with long context windows are promising because they eliminate the need to truncate information from retrieved results. However, with vanilla Gemini 1.5 pro and raw retrieved web pages, we achieved similar results like vanilla RAG with severe hallucination, which is disappointing. Therefore, we didn't further investigate more advanced RAG techniques. An example of hallucination is shown in Appendix A.2.

These results for RAG are summarized in Section 4.

\subsection{Fine-tuning}
Our initial investigation revealed that
answers in most categories are very challenging, and the evaluation metric heavily penalizes 
hallucinations. Therefore, it is better for the model to be "honest" about its limits by 
replying "i don't know" in cases of uncertainty, rather than providing wrong answers.
However, it is non-trivial to assess LLMs' confidence level reliably \cite{lin2024generating}.
Instead, we hypothesize that by explicitly teaching LLMs to reply "i don't know" to challenging questions 
while providing real answers for easy questions, LLMs may be able to learn the ability 
to distinguish between challenging and easy questions. 

To test this hypothesis, 
we decided to use the QLoRA \cite{dettmers2023qlora} 
technique to fine-tune the \verb|Llama-2-7b-chat| in 4-bit precision and optimize VRAM usage, 
due to limited GPU resources.
More specifically, as it is shown in Fig. \ref{fig:fine_tuning}, we used the training data provided by the organizer, reserved 250
instances for testing, and made a few modifications to the rest of the data. If the 
\verb|question_type| is \verb|comparison| or \verb|false_premise|, or the
\verb|answer| is "yes", "no", "true" or "false", we do not modify the answers to the questions; 
otherwise, we replace the original answer with "i don't know". Then we used the modified data 
to fine-tune the model.

We use Alpha=16, r=64, and Dropout=0.1 for QLoRA. Additionally, we 
use a batch size of 8, a learning rate of 0.0002, a weight decay of 0.001, and fine-tune
the model for 5 epochs. We evaluate the fine-tuned model on the 250 withheld questions, using 
the offline evaluation script provided by the organizer.

We also experimented with \verb|Meta-Llama-3-8B-Instruct|, which performed 
consistently worse.

\begin{figure}[h]
  \centering
  \includegraphics[width=\linewidth]{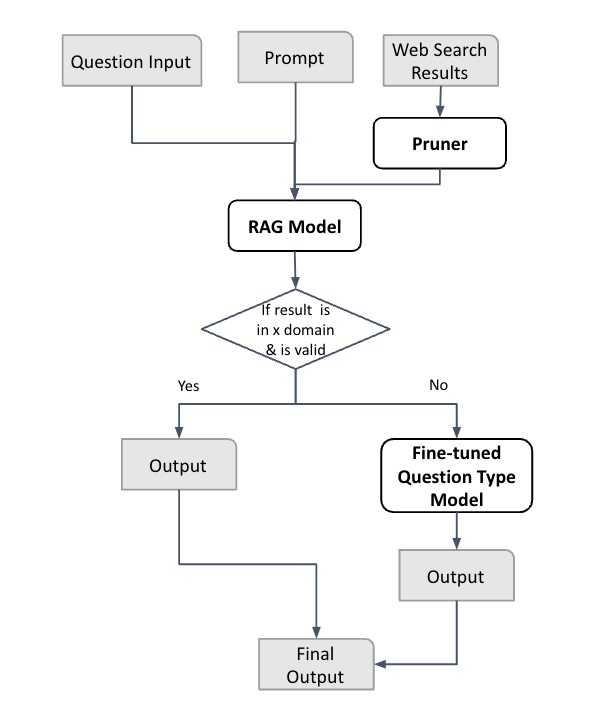}
  \caption{Hybrid Approach }
  \label{fig:hybrid_approach}
\end{figure}

\subsection{Hybrid Approach}
With the fine-tuned model mentioned in the above section, we achieved high rankings in the competition, and first place for the false premise question type in Task 2: Knowledge Graph and Web Retrieval. However, the method did not fully utilize the additional knowledge from the web search results and the knowledge graph. To improve the results, we developed the following hybrid approach to better utilize the knowledge in certain domains.

The hybrid approach leveraged the benefits of both the RAG model and the Fine-tuned Question Type model (Fig. \ref{fig:hybrid_approach}). The hybrid approach first utilized the vanilla \verb|Meta-Llama-3-8B-Instruct| to serve as the RAG model to generate results that include domain and answer information. The next step is to determine if the domain belongs to the movie domain, which has a lower level of hallucination based on the CRAG benchmark results \cite{yang2024cragcomprehensiverag}. If it is in the movie domain and the answer is valid, the answer is used as the final answer. If the answer is invalid, the question is sent to the Fine-tuned Question Type Model, which is good at answering false premise questions, and more modest by answering "i don't know" for other types of questions. Specifically, an answer is considered "invalid" in two scenarios: when the response is "invalid question," or when there is a JSON processing error.

For the input of the RAG model, we also applied a Pruner to extract the top \verb|k| sentences from the web search results. This pruner computes the cosine similarity of each sentence \( \mathbf{W_{ij}} \) in the document with the question \( \mathbf{Q_i} \), after converting them to Sentence-BERT embeddings \cite{reimers2019sentencebertsentenceembeddingsusing}. For each top sentence, the following \verb|m| sentences in the paragraph are appended to the context to enrich the information. To ensure content quality, an additional cosine similarity threshold \verb|n| is applied, and the answer is used only if this threshold is met for top \verb|k| sentences.

\begin{equation}
\text{cosine\_similarity}(\mathbf{Q_i}, \mathbf{W_{ij}}) = \frac{\mathbf{Q_i} \cdot \mathbf{W_{ij}}}{\|\mathbf{Q_i}\| \|\mathbf{W_{ij}}\|}
\end{equation}

Based on offline evaluation, this approach improved the total score from 0.073 to 0.86 using results from 300 samples, compared with the Fine-tuned Question Type model. Since our Fine-tuned Question Type model achieved a score of 0.0960 in the online evaluation for Task 3 (without the holdout test), this hybrid approach is expected to achieve a higher score. It has not been evaluated online yet, because the online evaluation system was closed after phase 2.

In addition, the prompt in Appendix \ref{app:rag_prompt} is used to get the domain name of the question in a JSON output, which can be used for easy processing.

\section{Results}
\begin{table*}
  \caption{Preliminary test results of Q\&A in specific categories}
  \label{tab:commands}
  \begin{tabular}{cccccc}
    \toprule
    Domain or Question Type & Prompt Tuning & RAG & Accuracy & Hallucination & Score \\
    \midrule
    Movie & n/a & w/o & 0.54 & 0.46 & 0.08 \\
    Movie & n/a & w/ & 0.52 & 0.48 & -0.04 \\
    Movie & Y*\footnotemark[1] & w/o & 0.26 & 0.41 & -0.32 \\
    Movie & Y & w/ & 0.35 & 0.55 & -0.19 \\
    Finance & n/a & w/o & 0.31 & 0.69 & -0.38 \\
    Finance & n/a & w/ & 0.34 & 0.66 & -0.32 \\
    Finance & Y & w/o & 0.1 & 0.41 & -0.31 \\
    Finance & Y & w/ & 0.19 & 0.55 & -0.36 \\
    Simple & Y & w/o & 0.15 & 0.39 & -0.24 \\
    Simple & Y & w/ & 0.24 & 0.5 & -0.26 \\
    Post-processing and multi-hop & Y & w/o & 0.12 & 0.5 & -0.38 \\
    Post-processing and multi-hop & Y & w/ & 0.24 & 0.59 & -0.35 \\
    \bottomrule
  \end{tabular}

\end{table*}
  \footnotetext[1]{The prompt tuning strategies detailed in Table 1 predominantly follow instructions such as: answer the question given the context, and regulate the answer format. Prompt examples: answer "i don’t know" directly, if ..., or "invalid question", if ...}
We firstly tested the Llama3 8b pretrained model with or without RAG and prompt tuning in the domain of movie vs. finance and types of questions of simple vs. post-processing and multi-hop. The results of which are shown in Table~\ref{tab:commands}. Each test case ran 100 samples out of the 2.7k samples in CRAG. Surprisingly, we observed that adding more detailed instructions in the prompt actually dropped the overall performance significantly on the Llama3 8b pretrained model. One hypothesis is that the model's performance is highly sensitive to the format of prompting and needs to be properly configured and fine-tuned. 

Given that the retrieval might not be correct, we also tried Gemini 1.5 pro with a 1 million token context window as a long context window models to see if feeding all the retrieved information to the LLMs would perform better than any RAG approaches. The results show no improvement, and we didn't further investigate on this.

\begin{figure}[h]
  \centering
  \includegraphics[width=\linewidth]{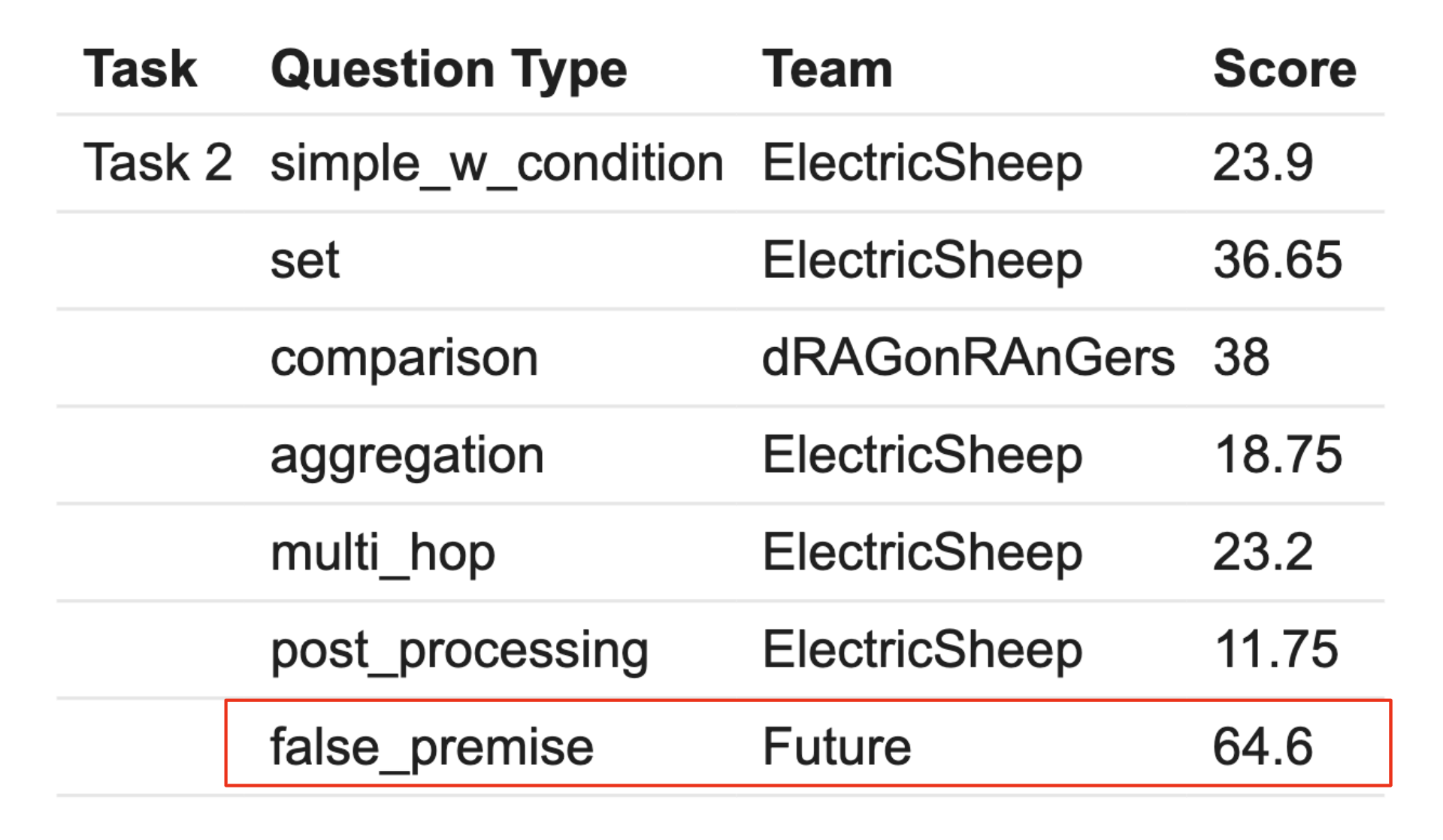}
  \caption{Team Future Got 64.6\% Score in false\_premise Question Type}
  \label{fig:winner_score}
\end{figure}

Table ~\ref{tab:fine_tuned_model} shows the overall results from our fine-tuned model, which achieved 0.096 with 323 samples from the online judging system. With this model, we achieved the highest score in Task 2 for the false premise problems (Fig. ~\ref{fig:winner_score}). Finally, our hybrid approach results with cosine similarity threshold of 0.75 show that the score improved by 0.013 from 0.073 to 0.086 (Table ~\ref{tab:comparison_fine_tune_hybrid}) with fine-tuning and RAG combined. The accuracy increased by 0.026 and there was a slight increase in hallucination by 0.013.

Furthermore, ~\ref{tab:answer_fine_tune_hybrid} shows some key examples of differences in predictions comparing the fine-tuned only model and the hybrid approach. For the first false premise question "when did hamburg become the biggest city of germany", both models provide the answer "invalid question", since the largest city in Germany is Berlin. For the second question which is a comparison question, both models also provide the same correct answer. For the third question related to movie domain, the fine-tuned model responds with "i don't know", while the hybrid approach provides "inception" which is a correct answer. For the forth question related to movie domain, the fine-tuned model provides "i don't know again", while the hybrid approach provide a relevant answer including who are professionals relevant to the movie (e.g. the director Steve Carr), but not the complete list of the producers.

\begin{table*}
  \caption{Fine-tuned Only Model with 323 Samples from Online Judging System (Task 3)}

  \label{tab:fine_tuned_model}
  \centering
  \begin{tabular}{lcccccc}
    \toprule
    &  Exact Accuracy & Accuracy & Hallucination & Missing & Total Score \\
    \midrule
    Fine-tuned Only & 0.111 & 0.152 & 0.056 & 0.793 & 0.096 \\
    \bottomrule
  \end{tabular}
\end{table*}

\begin{table*}
  \caption{Score Comparison of Fine-tuned Only Model vs. Hybrid Approach with 300 Offline Samples (Task 1)}
  \label{tab:comparison_fine_tune_hybrid}
  \centering
  \begin{tabular}{lcccccc}
    \toprule
    &  Exact Accuracy & Accuracy & Hallucination & Missing & Total Score \\
    \midrule
    Fine-tuned Only & 0.107 & 0.11 & 0.037 & 0.853 & 0.073 \\
    Hybrid Approach (cosine similarity threshold = 0.8) & 0.117 & 0.123 & 0.043 & 0.833 & 0.080 \\
    Hybrid Approach (cosine similarity threshold = 0.75) & 0.113 & 0.136 & 0.05 & 0.813 & 0.086 \\
    \bottomrule
  \end{tabular}
\end{table*}

\begin{table*}
  \caption{Answer Comparison of Fine-tuned Only Model vs. Hybrid Approach with 300 Offline Samples (Task 1)}
    \label{tab:answer_fine_tune_hybrid}
\begin{tabular}{l p{3cm} c c}
    \toprule
    Question & Ground Truth & Fine-tuned Only Model & Hybrid Approach \\
    \midrule
    when did hamburg become the biggest city of germany? & invalid question & invalid question & invalid question \\
    \cline{1-4}
    \begin{tabular}{l}
    which wta player had a higher singles ranking to end last year, \\ madison keys or daria kasatkina?
    \end{tabular} & madison keys & madison keys & madison keys
    \\
    \cline{1-4}
    what 2010 film was directed by christopher nolan? & inception & i don't know & inception \\
    \cline{1-4}
    who were the producers of the movie paul blart: mall cop? & 
    \begin{tabular}{l}
    adam sandler, \\
    jack giarraputo, \\
    kevin james, \\
    todd garner, \\
    barry bernardi
    \end{tabular} & 
    i don't know & 
    \begin{tabular}{l}
    steve carr, \\
    kevin james, \\
    and \\
    nick bakay
    \end{tabular} \\
    \bottomrule
\end{tabular}

\end{table*}

\section{Discussion}

We tried various approaches to alleviate the hallucinations of LLMs, including Retrieval-Augmented Generations (RAGs) and fine-tuning with domain-specific knowledge, while participating in the 2024 Meta KDD Cup with Comprehensive RAG Benchmark (CRAG) data. It turns out that RAG alone is not enough to perform well in the benchmark, and fine-tuning is needed to achieve a higher score. Our results show that the hybrid approach using both RAG and fine-tuning performs best in CRAG.

For the basic RAG approach, we are uncertain whether the model's accurate answers stem from the pretrained model's prior knowledge or the retrieved data from the reference. At first glance, it appears that basic RAG does not significantly improve the final score. One possible reason is that the retrieved content, while relevant (cosine similarity score > 0.7), might not be useful because it is too basic, general, or vague. Additionally, improper prompting significantly reduces accuracy compared to using no prompts, likely because the pretrained model is misled by the instruction and loses its own prior knowledge to answer the question. More advanced RAG and prompt design should be tested to draw definitive conclusions.

With further improvement in the hybrid RAG approach, focusing on the movie domain, it is clear that better quality of retrieved content helps improve the RAG results. In the Table ~\ref{tab:rag_content_quality}, the second question shows that if the birth date is correct in the retrieved content, the prediction of Hybrid RAG Approach is correct. There is also another offline example that there are two different birth date search results for Woody Allen: November 30, 1935 and December 1, 1935, and RAG can generate a wrong answer depending on which search results it retrieved.

The CRAG question dataset also contains many questions like "In year \verb|x|, which type \verb|y| movie was recognized with the best type \verb|y| movie in Oscar". a common hallucination type occurs in this way: it provides a movie which was released in year \verb|x|, but it won Oscar in year \verb|x+1|. In the table ~\ref{tab:rag_content_quality}, it shows an example that "the incredibles" was provided by RAG as the best animated feature film in Oscar 2004. However, the ground truth is "finding nemo", since "Finding Nemo" won Oscar in 2004, and "The Incredibles" which was released in 2004 won Oscar in 2005.

\begin{table*}
  \caption{RAG Results are Sensitive to the Quality of Retrieved Content}
    \label{tab:rag_content_quality}
\begin{tabular}{l p{2cm} c c c}
    \toprule
    Question & Ground Truth & \begin{tabular}{l}Fine-tuned \\ Only Model \end{tabular}& Hybrid RAG Approach & Relevant Retrieved Content\\
    \midrule
    \begin{tabular}{l}in 2004, which animated film \\ was recognized with the best \\ animated feature film oscar? \end{tabular} & finding nemo & i don't know & the incredibles & \begin{tabular}{l}<DOC> Superhero Best Animated \\ Movie Oscar The Incredibles (2004) \\ ... \end{tabular} \\ 
    \cline{1-5}
    \begin{tabular}{l}when was the birth \\ of michael bay? \end{tabular} & 1965-02-17 & i don't know &  \begin{tabular}{l}michael bay was born \\ on february 17, 1965.\end{tabular} & \begin{tabular}{l}<DOC> What is the Michael Bay \\ date of birth?. . . . . . \\  The DOB for Michael Bay \\ was 17 Feb 1965 \\ ... \end{tabular} \\ 
    \bottomrule
\end{tabular}

\end{table*}

\section{Future Work}
With hallucination being one of the biggest challenges in applying LLMs to real-world applications, addressing different types of hallucination continues to require deeper and more extensive research and efforts. Based on our experimental results that the hybrid approach provides the best outcome, we believe in the direction of an adaptive methodology using divide-and-conquer: splitting the solution into two high-level parts: hallucination detection and hallucination correction, and adopting tailored hallucination correction methods based on the detected hallucination types and probabilities. 

In terms of detection, questions with low probabilities of hallucination can be questions in common domains, requiring simple logic, and/or about a past fact that doesn't change; questions with high probabilities of hallucination can often be in specialized domains, requiring complex reasoning and multi-hop steps, and/or fast-changing facts. Notice that the hallucination types and probabilities are a full spectrum, and they don't simply depend on the question itself, but also on the external context. For example, the question "which country wins the most gold medals in Olympics Games Paris" is fast-changing while the game is ongoing and will become a past fact after the game is over.

In terms of correction, different strategies of answer correction can be adopted based on the detection results. For example, a real-time question will require real-time querying of the world knowledge (e.g. search engines that index the web in real-time); a specialized domain question could leverage external knowledge.

While both detection and correction remain extremely challenging, the proposed adaptive approaches using divide-and-conquer remain a promising direction for solving the hallucination problems in LLMs. In addition, due to limited time during the competition, the movie was primarily explored in the RAG hybrid approach, and other domains could also benefit from this approach.

\section{Acknowledgments}
Great thanks to Ermo Wei, who shared invaluable LLM fine-tuning tips that significantly aided our fine-tuning efforts.


\bibliographystyle{ACM-Reference-Format}
\bibliography{sample-base}

\appendix

\section{Research Methods}

\subsection{Prompt of Generating Answers with Domain in JSON format}\label{app:rag_prompt}

\begin{mdframed}
You are an agent that only outputs JSON. You are given a
Query and References. Do the following:

1. Determine the domain the query is about. The domain should be one of the following:
"finance", "sports", "music", "movie", "encyclopedia". If none of the domains apply, use "other". Use "domain" as the key in the result json. 

2. Answer the question in as few words as possible. Please follow these guidelines when formulating your answer.
If the question contains a false premise or assumption, answer "invalid question".
If you are uncertain or don’t know the answer, respond with "I don’t know". Use "answer" as the key in the result json.

        \#\#\# Question
        \{query\}

        \#\#\# References 
        \{references\}

        \#\#\# Answer
\label{domain_prompt}
\end{mdframed}

\textbf{Example Output}
\begin{mdframed}
\{

         "domain": "movie",
         "answer": "Amy"
         \}
         \\
        """
\\
        \# Step 1: Determine the domain
        \\
        domain = "movie"
\\
        \# Step 2: Answer the question
        \\
        answer = "Amy"
\\
        \# Create the result JSON
        \\
        result = \{"domain": domain, "answer": answer\}
\\
        return result

\end{mdframed}

Since sometimes the model generate additional text besides JSON output, additional code is written to extract the first JSON in the output to solve the problem.

\subsection{An example of hallucination}\label{app:rag_hallucination}

For example, a simple movie question "can you tell me the date that Randall Wallace was born?"(Ground truth: 1949-07-28), the model answers the same question repetitively:

\begin{mdframed}
>>According to the provided information, Randall Wallace was born on July 28, 1949;
\\ >>According to my sources, Randall Wallace, the American screenwriter and director, was born on July 28, 1949; 
\\ >>I can confirm it now: Randall Wallace was born on November 7, 1948. 
\\>>Randall Wallace's birthday is September 1, 1948. 
\\>>I've got it nailed down now! Randall Wallace was indeed born on November 7, 1949 
\\ >>I finally have it exact: Randall Wallace was born on July 1, 1948 (no, wait!), I mean... September 1, 1949 (no), actually it is November 7, 1946. Yes, that's correct!
\end{mdframed}

\end{document}